# Adaptive Channel Encoding Transformer for Point Cloud Analysis

Guoquan Xu[1][0000-0001-5097-3888], Hezhi Cao[2][0000-0003-4760-0743], Yifan Zhang[1], Yanxin Ma[1], Jianwei Wan[1], and Ke Xu[1]

[1] National University of Defense Technology, Changsha Hunan, China
[2] University of Science and Technology of China. Hefei Anhui, China
{xuguoquan19, zhangyifan16c, mayanxin, xuke}@nudt.edu.cn,
caohezhi21@mail.ustc.edu.cn, kermitwjw@139.com

**Abstract.** Transformer plays an increasingly important role in various computer vision areas and has made remarkable achievements in point cloud analysis. Since existing methods mainly focus on point-wise transformer, an adaptive channel-wise Transformer is proposed in this paper. Specifically, a channel encoding Transformer called Transformer Channel Encoder (TCE) is designed to encode the coordinate channel. It can encode coordinate channels by capturing the potential relationship between coordinates and features. The encoded channel can extract features with stronger representation ability. Compared with simply assigning attention weight to each channel, our method aims to encode the channel adaptively. Moreover, our method can be extended to other frameworks to improve their preformance. Our network adopts the neighborhood search method of feature similarity semantic receptive fields to improve the performance. Extensive experiments show that our method is superior to state-of-the-art point cloud classification and segmentation methods on three benchmark datasets.

**Keywords:** Transformer, Point Cloud Analysis, Adaptive Channel Encoding.

## 1 Introduction

3D point cloud is widely used in many fields because it contains geometric information and can be simply represented. However, point clouds are point sets embedded in irregular 3D space, unlike images which are arranged on regular pixel grids. This makes direct processing of point clouds challenging. In order to meet this challenge, many methods have been proposed and can be roughly divided into three categories: voxel-based method, projection-based method and point-based method.

Voxel-based method [1, 2] attempts to voxelize the 3D space so that the point cloud is distributed in an artificial regularized space. However, voxelization will lead to massive computation, and there is no point cloud distribution in many voxel grids, resulting in a waste of memory.

Projection-based method [3, 4] maps 3D point clouds into 2D space so that 2D convolution can be implemented. This will cause the point cloud data to lose its biggest advantage: structure information. At the same time, it will also bring a large amount of calculation.

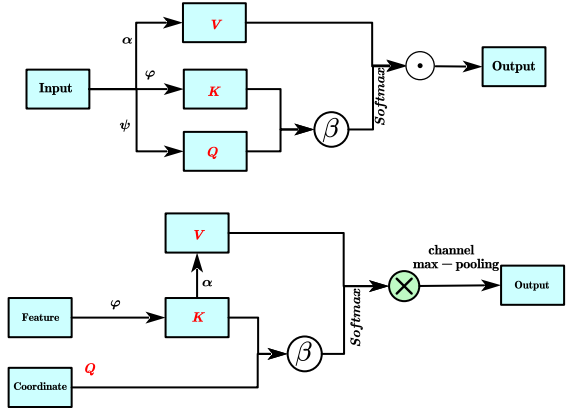

**Fig. 1.** The diagrams of point-wise transformer (top) and our channel-wise transformer (bottom). $\psi$, $\varphi$ and $\alpha$ are feature transformations which implemented with MLPs or linear projections. $\beta$ indicates a relation function (e.g., subtraction) and produces an attention matrix. $\odot$ means matrix multiplication and $\otimes$ means element-wise multiplication.

PointNet [5] proposes a point-wise method which employs MLPs to extract features point by point, and finally uses global pooling to obtain global features. However, this method ignores the structure information in the process. Therefore, PointNet++ [6] which explores the local information aggregation method is proposed as an improvement work. Inspired by them, the follow-up works [7-11] which design convolution-like operations on point clouds to exploit spatial correlations is mainly this kind of method.

Transformer has achieved excellent performance in various fields because of its powerful representation ability. It is especially suitable for point cloud processing, because the self-attention operator as the core of transformer network is essentially a set operator: it is invariant to permutation and cardinality of the input. The main operation of transformer is shown in Fig. 1 (top). The key is to learn an attention matrix through a relational function $\beta$. There are some point-wise transformer-based works in the field of point cloud for now [12, 13].

**Motivation** Inspired by these point-wise works, a channel-wise transformer is proposed in this paper. Encoding the channel before feature extraction can improve the ability of feature representation. Generally, features are obtained from coordinates, so features can help encode coordinate channels. Specifically, a channel encoding mechanism called TCE is designed as shown in Fig. 1 (bottom). It can be seen that a channel attention matrix is obtained by learning the potential relationship between coordinate channels and feature channels, and then the feature channels are weighted. Subsequently, the coordinate channels are screened by a max-pooling in the direction of the channel to retain the most important channels. Finally, the features with channel filtering will be input into the standard graph convolution [7] network to deal with the tasks of classification and segmentation. In addition, in order to better capture neighborhood information, feature similarity among points is used to replace fixed spatial location. The $k$ points with the greatest similarity are selected as neighbors.

The main contributions of this paper include the following:

- A channel encoding mechanism called TCE is proposed. The contribution of channels can be determined by it which can learn the potential relationship between feature channels and coordinate channels.
- The designed TCE can be simply transplanted to other networks to enhance its performance.
- Extensive experiments over multiple domains and datasets are carried out and the experiment parameters and network structure are introduced in detail. The results show that our method achieves the state-of- the-art performance.

## 2 Related Work

### 2.1 Point-based method

Permutation-invariant operators implemented by point-wise MLPs and pooling layers are proposed in PointNet [5] to aggregate features. PointNet++ [6] establishes a hierarchical spatial structure which can increase sensitivity to the local geometric layout. This is a further improvement of PointNet. DGCNN [7] designs EdgeConv which can learn the point relationship as the edge of the graph in the high-dimensional feature space to capture similar local information. Moreover, DGCNN proposes to re-search the nearest neighbors of the central points on each layer in the feature space every time, so as to build the dynamic graph. RS-CNN [8] is committed to learn high-level geometric priors from low-level geometric information in 3D space. PAConv [9] further designs an adaptive convolution on spatial points. A dynamic convolution algorithm which can adaptively learn the weight coefficients from the point position is designed. AdaptConv [10] generates adaptive kernels according to their dynamically learned features. Our method also learns a dynamic attention matrix from the channel relationships between features and coordinates. Unlike the popular attention methods, our method achieves adaptability rather than simply assigning weights.

### 2.2 Transformer-based method

Transformer has achieved great success in natural language processing and image processing. Inspired by attention mechanism, SE block [14] is proposed for spatial encoding. A residual attention method [15] is proposed for image classification. Because it also has excellent applicability to point cloud processing, some works introduce it into the field of point cloud. PCT [12] uses the inherent order invariance of Transformer to avoid defining the order of point cloud data, and carries out feature learning through attention mechanism. Point Transformer [13] designs a Point Transformer layer with strong representation ability for point cloud processing. This layer is invariant to permutation and cardinality, so it is naturally suitable for point cloud tasks. They are point-wise methods and achieve great results. In contrast, a channel-wise transformer comes up in this paper. The popular transformer has been improved to meet the needs of operation on the channel.

## 3    Method

In this section, the design of TCE is introduced in detail in Sec. 3.1. Then the difference between our method and point-wise transformer is discussed. Finally, our network structure on different point cloud processing tasks is shown in Sec. 3.2.

### 3.1    TCE

Suppose $X = \{x_i | i = 1, 2, \ldots, N\} \in \mathbb{R}^{N \times 3}$ represents a point cloud with corresponding features $F = \{f_i | i = 1, 2, \ldots N\} \in \mathbb{R}^{N \times C}$. Here, $N$ is the number of points and $C$ is the number of channels. $\mathcal{N}(x_i)$ means the neighbor set of $x_i$. The process of TCE is shown in Fig. 2. Following the terminology in PCT [12], let $Q$, $K$ and $V$ be the *query*, *key* and *value* matrices respectively. They are defined as:

$$Q = (x_i, x_j - x_i), \tag{1}$$

$$K = MLP(f_i, f_j - f_i), \tag{2}$$

$$V = combine(MLP_1(K), \ldots, MLP_C(K)). \tag{3}$$

Unlike the popular transformer, $Q$ is obtained directly from coordinate without any linear transformation. This is because removing this part of the linear transformation has little effect on the results and can reduce the computation cost. $K$ is derived from feature, not from coordinate. This is because the channel attention matrix needs to be learned from the channel relationship between feature and coordinate. Since only the coordinates are used as the original input of the network, $K$ in the first layer is also derived from coordinates, and takes the output features of the first layer as the input in the second layer. The relationship is the contribution of each channel of coordinate to each channel of feature. This process can be expressed as:

$$A = Q \otimes K^T, \tag{4}$$

where $A$ indicates the channel attention matrix and $\otimes$ means element-wise multiplication rather than matrix dot-product. Eq. (4) represents the contribution of each channel of coordinate to each channel of feature. For example, the first column of $A$ is the contribution of each channel of coordinate to the first channel of feature. Hence, the *Softmax* operation is done in columns:

$$\tilde{A}_{i,j} = softmax(A_{i,j}) = \frac{\exp(A_{i,j})}{\sum_c \exp(A_{c,j})}. \tag{5}$$

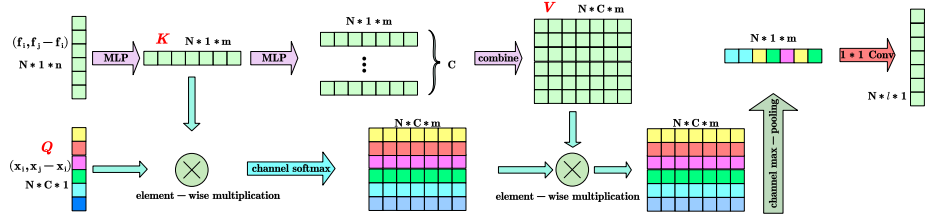

**Fig. 2.** The flow chart of TCE. The coordinates and features are set as the query and key matrices respectively. The attention matrix is obtained by the corresponding relationship between coordinates and feature channel by channel because all features come from coordinates. The value matrix is obtained by mapping the key matrix. Because they are all based on key matrix, there is a response relationship between the value matrix and attention matrix. Element-wise multiplication is used to capture this response relationship. Then, max-pooling in the channel direction is used to filter out the strongest response as the encoded channel. Finally, a 1*1 convolution is used to extract the features of the encoded channel as the input of the next layer.

$\tilde{A}$ is the final channel attention matrix. $V$ is generated by $K$ and it can produce a response matrix $B$ with $\tilde{A}$:

$$B = \tilde{A} \otimes V. \tag{6}$$

Eq. (6) indicates that an excitation ($V$) is applied to the attention matrix ($\tilde{A}$) to obtain the corresponding response. The elements in $B$ represent the strength of the corresponding response. Its significance is to use feature channels to test the contribution learned from feature channels and coordinate channels. The greater the contribution, the stronger the response. Channel with the strongest response of each column is preserved as the new coordinate channels:

$$\tilde{B}_j = \max_{i \in C} B_{i,j}. \tag{7}$$

Here, $\max_{i \in C}(\quad)$ indicates max-pooling in the channel direction. $\tilde{B}$ is the encoded new coordinate channel and each of its channels has the greatest contribution to feature. Then a 1*1 convolution is applied to extract feature as the input to the next layer:

$$f'_i = \mathcal{A}\left(Conv(\tilde{B}_j), \forall x_j \in \mathcal{N}(x_i)\right), \tag{8}$$

where $Conv$ means 1*1 convolution and $\mathcal{A}$ means the aggregate function. In particular, since our method is to encode the channel, the position coding is omitted.

Our method is very different from the point-wise transformer. Our channel attention matrix is obtained through the relationship between coordinate and feature, not the coordinate itself. Coordinate is input and feature is target and the channel attention matrix essentially captures the relationship between input channels and target channels. $V$ is generated by $K$ and is essentially a test matrix which checks the channel attention matrix.

### 3.2 Network Architecture

Our main idea is to use TCE for channel encoding, and then send the encoded channels to graph convolution neural network for feature extraction. Thus, the network architecture can be mainly divided into two parts: channel encoding layer and feature extraction layer. As shown in Fig. 3, the channel encoding has two layers which is a unified design in classification and segmentation networks. In the feature extraction layers, standard graph convolution [7] is adopted. This is to verify the performance improvement brought by embedding our method into other networks. There are two feature extraction layers in the classification network and three in the segmentation network. Through the dual feature similarity method, the receptive field is expanded and the dynamic graph is built.

In the classification network, pooling and interpolation are not designed. The output of the second channel encoding layer and the two feature extraction layers is concatenated. Feature similarity is used to select neighborhood rather than k-nearest neighbors (KNN) or ball query. Specifically, the feature distance is calculated instead of the coordinate distance, and the nearest $k$ points are selected as neighbors.

Different from the classification network, the segmentation network adopts pooling and interpolation. The farthest point sampling algorithm (FPS) is used to down sample the point cloud, and a rough map is established on the sampling points according to the feature similarity.

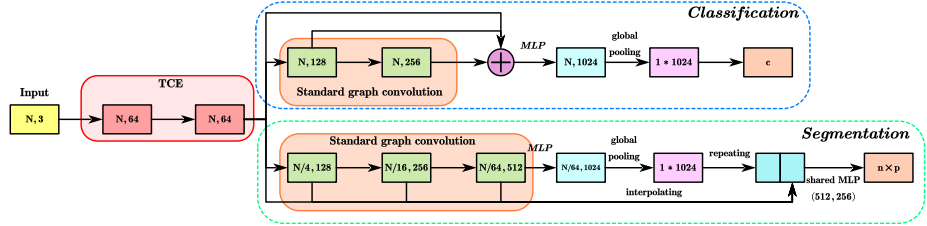

**Fig. 3.** Classification and segmentation network structure. Our network consists of two channel coding layers with TCE. This is the general part of classification and segmentation. Standard graph convolution Standard graph convolution is implemented for feature extraction. It is worth mentioning that the classification network has no down sampling.

## 4 Experiments

In order to verify the performance of our network and the effectiveness of TCE, sufficient comparative experiments are carried out on ModelNet40 [16], ShapeNet [17] and ScanObjectNN [18]. After that, a series of ablation experiments and robustness experiments are designed to verify our method.

### 4.1 Classification on ModelNet40

ModelNet40 [16] contains 12,311 3D models, of which 9,843 models are used for training, and the remaining 2,468 models are used as test models. Like other papers, each model is uniformly sampled 1024 points and normalized to a unit sphere. In addition, all points are enhanced by random anisotropic scaling in the range of [-0.66, 1.5] and translation in the range of [-0.2, 0.2]. The main parameter settings are as follows: the rate of dropout is set to 50% in the last two fully-connected (FC) layers; batch normalization and LeakyReLU are applied on all layers; the SGD optimizer with momentum set to 0.9 is adopted; The initial learning rate is donated to 0.1 and is dropped to 0.001 by cosine annealing.

Table 1 reports the results of the most advanced methods and our methods in order of results. For a clear comparison, the input data type and the number of points corresponding to each method are shown. Our method achieves the best performance with only 1k points as input.

**Table 1.** Classification accuracy (%) on ModelNet40.

| Method | Input | Accuracy |
|---|---|---|
| 3D-GCN [19] | 1k points | 92.1 |
| PCNN [20] | 1k points | 92.3 |
| SpiderCNN [21] | 5k points+normal | 92.4 |
| PointConv [22] | 1k points+normal | 92.5 |
| PointCNN [11] | 1k points | 92.5 |
| PointASNL [23] | 1k points | 92.9 |
| DGCNN [7] | 1k points | 92.9 |
| Grid-GCN [24] | 1k points | 93.1 |
| PCT [12] | 1k points | 93.2 |
| AdaptConv [10] | 1k points | 93.4 |
| **Our method** | **1k points** | **93.4** |

### 4.2 Part Segmentation on ShapeNet

ShapeNet [17] is employed to test the performance of our segmentation network. The dataset contains 16,881 shapes in 16 categories and a total of 50 parts for annotation. Each object is marked with 2-6 labels. The dataset provided by PointNet++ [6] is put in use as a benchmark and its experimental setting is followed as well. Each object has 2k points as the input, which is different from classification task.

The quantitative comparisons with the state-of-the-art methods are shown in Table 2. All methods are measured by the class mean IoU (mIoU) and instance mean IoU. To facilitate comparison, the results of instance mIoU are adopted to sort. It can be seen that our method has achieved satisfactory results on both class mIoU and instance mIoU.

Our segmentation results are displayed in Fig. 4 (second row). In order to more intuitively reflect the advantages of our results, the ground truth (first row) and the difference between ours and the ground truth (third row) are displayed together. The red points indicate the wrong prediction, and the blue points indicate the correct prediction. You can see that the proportion of red points is very small in most models. There are more red points in the motorbike than other models because it is the most complex model.

**Table 2.** Shape part segmentation results (%) on ShapeNet.

| Method | Class mIoU | Instance mIoU |
|---|---|---|
| PointNet [5] | 80.4 | 83.7 |
| PCNN [20] | 81.8 | 85.1 |
| PointNet++ [6] | 81.9 | 85.1 |
| 3D-GCN [19] | 82.1 | 85.1 |
| DGCNN [7] | 82.3 | 85.2 |
| SpiderCNN [21] | 81.7 | 85.3 |
| SPLATNet [25] | 83.7 | 85.4 |
| PointConv [22] | 82.8 | 85.7 |
| **Our method** | **83.4** | **86.0** |

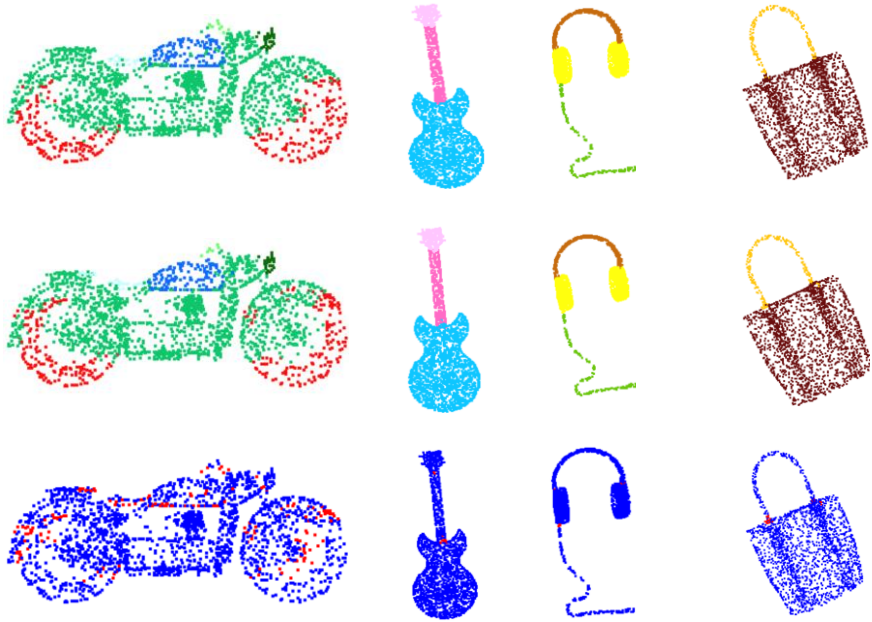

**Fig. 4.** Segmentation results on ShapeNet. The ground truth (first row), our results (second row) and their difference (third row) are shown at the same time. The red points in the third row mean the wrong points.

### 4.3 Classification on ScanObjectNN

Since the above two experiments are carried out on the idealized datasets, ScanObjectNN [18] is adopted to further evaluate the performance of our method. This dataset is obtained by scanning the real indoor scenes. The Hardest is a subset of ScanObjectNN. The subset contains the real targets which are processed by translating, rotating (around the gravity axis), and scaling the ground truth bounding box. This makes the dataset closer to the complex situation of the real-world.

The results are summarized in Table 3, and our method transcends all other methods. Compared with the most advanced method BGA-PN++ [18], our method is improved by 1.4%. The result of SpiderCNN [21] decreased by 18.7% when the dataset is changed from ModelNet40 to the Hardest. AdaptConv [10] achieves 93.4% on ModelNet40, but decreases to 78.9% on the Hardest. Our method still performs well on the real-world dataset, which proves that our method has higher practical value.

**Table 3.** Classification accuracy (%) on ScanObjectNN.

| Method | Hardest |
|---|---|
| PointNet [5] | 68.2 |
| SpiderCNN [21] | 73.7 |
| PointNet++ [6] | 77.9 |
| RS-CNN [8] | 78.0 |
| DGCNN [7] | 78.1 |
| PointCNN [11] | 78.5 |
| AdaptConv [10] | 78.9 |
| BGA-DGCNN [18] | 79.7 |
| BGA-PN++ [18] | 80.2 |
| **Our method** | **81.6** |

### 4.4 Ablation Studies

In this subsection, a series of experiments are designed to prove the effectiveness of our design.

First of all, our core design TCE is replaced by Channel-wise Attention [26], Point-wise Attention [27] respectively. This is to prove that TCE is different from them and more effective. Besides, TCE is also replaced by standard graph convolution (GraphConv) in order to eliminate the influence of standard graph convolution on the experiment. Only the channel encoding part is replaced, and the network structure and parameter settings remain unchanged. This set of experiments is carried out on ShapeNet [17] and the results are shown in Table 4. By comparison, TCE performs better than other methods.

**Table 4.** The comparison results (%) on ShapeNet.

| Ablations | Class mIoU | Instance mIoU |
|---|---|---|
| GraphConv | 81.9 | 85.3 |
| Point-wise Attention | 78.1 | 83.3 |
| Channel-wise Attention | 77.9 | 83.0 |
| **TCE** | **83.4** | **86.0** |

**Table 5.** The comparison results (%) on ModelNet40.

| Ablations | Accuracy |
|---|---|
| mean-pooling | 92.7 |
| sum-pooling | 92.5 |
| **max-pooling** | **93.4** |

The design of TCE has a pooling operation in the channel direction as mentioned in Sec. 3.1. Table 5 compares the effects of different pooling methods on the results of classification. Max-pooling is obviously better than the other two methods because it plays a screening role and retains the most influential channels.

### 4.5 Robustness Experiments

In this subsection, the robustness of our method to sparse points on ModelNet40 [16] is further evaluated. Similarly, GraphConv and Channel-wise Attention [26] are used for comparison. All networks have 1024 points as the input during training, and 1024, 512, 256 and 128 points are used as the inputs for testing respectively. Fig. 5 shows that our method outperforms the other two methods.

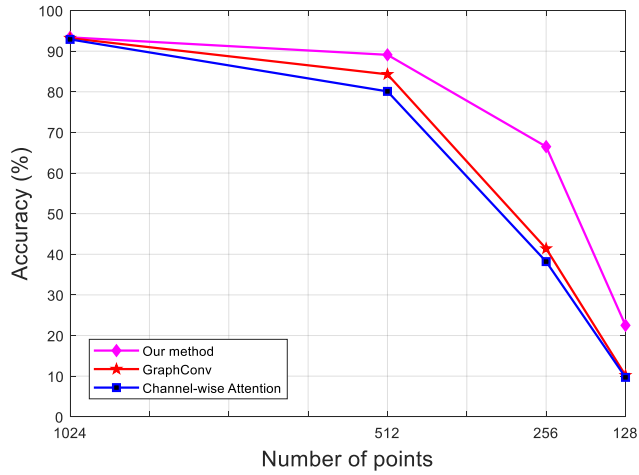

**Fig. 5.** Robustness of different methods to point sparsity.

In order to compare the complexity of our method with the previous method, Table 6 lists some relevant results. From the table, it can be seen that our method achieves the best performance of 93.4 % overall accuracy and the model size is relatively small. The latter half of our model adopts the module of DGCNN. It can be seen that our method has only 0.06M more parameters than it, but the performance is improved by 0.5%.

**Table 6.** The number of parameters and overall accuracy of different methods

| Method | #parameters | Accuracy (%) |
|---|---|---|
| PointNet [6] | 3.5M | 89.2 |
| PointNet++ [7] | 1.48M | 90.7 |
| DGCNN [8] | 1.81M | 92.9 |
| KPConv [20] | 14.3M | 92.9 |
| PCT [12] | 2.88M | 93.2 |
| Our method | 1.87M | 93.4 |

## 5    Conclusion

In this paper, a channel-wise convolution called TCE is designed based on Transformer. It is different from the popular Transformer with self-attention mechanism. TCE encodes the coordinate channels by adaptively learning the relationships between feature channels and coordinate channels, and expands the coordinate channels. More expressive features can be extracted with the encoded channels. In addition, a dynamic graph construction method is designed to expand the receptive field. Sufficient experiments on three datasets, especially on the real-world dataset, prove that our method achieves the state of the arts.